# Pruning Bayesian Networks for Efficient Computation

Michelle Baker and Terrance E. Boult
Department of Computer Science
Columbia University
New York, NY 10027
Baker@cs.columbia.edu, Tboult@cs.columbia.edu

## 1 Abstract

This paper analyzes the circumstances under which Bayesian networks can be pruned in order to reduce computational complexity without altering the computation for variables of interest. Given a problem instance which consists of a query and evidence for a set of nodes in the network, it is possible to delete portions of the network which do not participate in the computation for the query. Savings in computational complexity can be large when the original network is not singly connected.

Results analogous to those described in this paper have been derived before [Geiger, Verma, and Pearl 89, Shachter 88] but the implications for reducing complexity of the computations in Bayesian networks have not been stated explicitly. We show how a preprocessing step can be used to prune a Bayesian network prior to using standard algorithms to solve a given problem instance. We also show how our results can be used in a parallel distributed implementation in order to achieve greater savings. We define a minimal computationally equivalent subgraph of a Bayesian network. The algorithm developed in [Geiger, Verma, and Pearl 89] is modified to construct the subgraphs described in this paper with $O(e)$ complexity, where $e$ is the number of edges in the Bayesian network. Finally, we define a minimal computationally equivalent subgraph and prove that the subgraphs described are minimal.

## 2 Introduction

The computation of conditional probabilities for arbitrary discrete probability distributions is very efficient in singly connected Bayesian networks. However, in the general case the problem is NP-Complete [Cooper 89]. This paper examines how the specific query one is interested in and the evidence available can be taken advantage of in order to reduce computational complexity. If one is interested in determining values for a subset of the variables in a problem domain it is not necessary to propagate information along every path in the network. Thus the network can be pruned prior to carrying out the computation. Very large savings in computational complexity are possible when a multiply connected network can be reduced to a singly connected subgraph.

One of the implications of this paper is that it is not necessary that evidence be available in order for a Bayesian network to be pruned. Probably the simplest example in which enormous savings in computation are possible is the case in which you are only interested in knowing the value of a root (i.e., parentless) node of a network and there is no evidence available. In this case, the network can be pruned until only the single node that you are interested in remains. The example seems trivial because, when there is no evidence, we need to determine the prior probability of the nodes we are interested in and root nodes have their priors directly available. Nevertheless, techniques for pruning Bayesian networks that are based only on probabilistic independencies among variables would return the entire network when given this example.

The runtime construction of subgraphs of Bayesian networks in order to reduce computational complexity has only recently become a focus of research. [Wellman 88] developed methods for constructing qualitative Bayesian networks at various levels of abstraction. Another technique, more closely related to the work described in this paper, has used evidence to guide dynamic network construction. In his work on combining first order logic with probabilistic inference, [Breese 89] designed an algorithm for dynamic network construction that is based on evidence induced probabilistic independence. Using the semantics of *d-separation* he proves the algorithm correct in the sense that subgraphs generated do not introduce unwarranted assumptions of probabilistic independence.

The main contribution of this paper is to show that as part of the dynamic construction of a subgraph of a Bayesian network, leaf nodes without evidence can be recursively removed without altering the computation at nodes of interest. Following [Shachter 88], we will call these *barren nodes*. As was illustrated in the previous example, barren nodes need not be d-separated from the nodes of interest in order to be removed. Although similar results have been stated before [Geiger, Verma, and Pearl 89, Shachter 88], their implications for the runtime construction of subgraphs of Bayesian networks has not been clear. The results of the analysis in [Shachter 88] of the informational requirements for solution of a problem instance using an influence diagram is equivalent to the results described here for Bayesian networks. However, because the solution algorithm for influence diagrams is bound up with the graph reduction it is not immediately obvious how one would apply

Shachter's methods to Bayesian networks. Indeed, this has not been done in an implemented system. Alternatively, theoretical work in [Geiger, Verma, and Pearl 89] that analyzes the distinction between "sensitivity to parameter values" and sensitivity to variable instantiations" imply results described in this paper. However, because that work did not address the question of efficient computation and treated the two types of independence as separate issues with separate algorithms its full implications for the dynamic construction of Bayesian networks were hidden.

In the rest of this paper we will define formally what is meant by a computationally equivalent subgraph, prove that recursive pruning of leaf nodes without evidence does not violate computational equivalence and show how the algorithm developed by [Geiger, Verma, and Pearl 89] can be used to construct subgraphs described in this paper. Finally, we define formally what is meant by a minimal computationally equivalent subgraph and prove that the subgraphs described here are minimal. In the conclusion we discuss how our results apply additional savings in a parallel distributed implementation and discuss limitations of the method.

## 3 Computational equivalence v.s. d-separation

The specific problem addressed in this paper is that of finding the smallest subgraph of a Bayesian network that will correctly compute the conditional probability distributions for a subset of the variables in the network. Given a Bayesian network and a problem instance which consists of evidence for a set of variables and another set of variables whose values we wish to know, we would like to find the smallest subgraph (or set of subgraphs) of the network such that the computation for each of the variables of interest is unchanged.

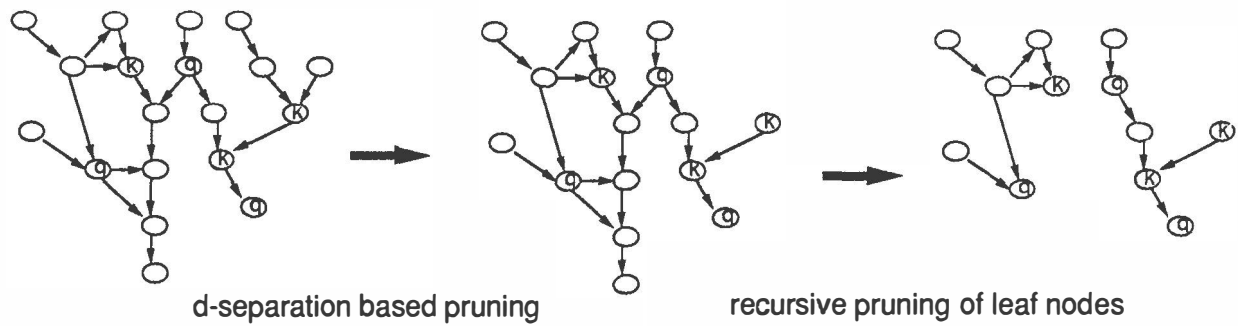

d-separation based pruning        recursive pruning of leaf nodes

**Figure 1:** Two computationally equivalent subgraphs for a problem instance

A natural approach to take in solving this problem is to prune away nodes that correspond to variables that are probabilistically independent of the variables of interest. Probabilistic independencies are represented graphically in Bayesian networks according to a semantics defined by *d-separation* [Pearl 88]. Evidence for variables that are known with certainty define a set that d-separates other pairs of sets of variables. Using d-separation for pruning amounts to finding the set of variables d-separated from the nodes of interest by the evidence. An algorithm that is linear in the number of edges in the underlying network has been designed to solve this problem [Geiger, Verma, and Pearl 89]. However, whereas an algorithm based on d-separation is sufficient to guarantee computational equivalence, there are cases in which nodes that are not d-separated from the nodes of interest can be removed without jeopardizing computational equivalence. In particular, barren nodes can be removed until either a node with evidence or a node of interest is encountered.

**DEFINITION:** *Let Q be a subset of the nodes in a Bayesian network. A subgraph of a Bayesian network is* **computationally equivalent** *to the network with respect to Q if, for each node q ∈ Q, the computation at that*

*node, including the computation for BEL(q), λ(q), and π(q) is identical in the subgraph to the computation at that node in the original network.*

Figure 1 illustrates two examples of computationally equivalent subgraphs. Nodes labeled *q* are in *Q*, i.e., they represent the variables whose values we are interested in knowing. Nodes labeled *k* are evidence, i.e., they represent variables which have known values. The subgraph on the left is constructed by an algorithm based entirely on d-separation. The one on the right is constructed by removing barren nodes as well as by removing all the nodes that are d-separated from the query nodes by the evidence.

The following theorem provides the basis for the claim that barren nodes can be removed from a Bayesian network without affecting the computation for a selected subset of nodes. Furthermore, as one would expect, all nodes that are d-separated by the evidence set from the nodes of interest can be removed. The subgraph that is constructed by this method may not be connected but there will be at most one graph for each node in *Q*.

**Theorem 2.1:** *Let Q denote the set of nodes whose values we are interested in and K be*

*the set nodes with known values. A subgraph, G, of a Bayesian network, D, is computationally equivalent to D with respect to a problem instance, (Q, K), if it is constructed by (1) removing all nodes that are d-separated from Q by K, (2) removing barren nodes until either a node in Q or a node in K is found, and (3) removing all edges that are not incident on two nodes in G.*

**Proof:**

1. From the definition of d-separation we know that if two sets of nodes, $Q$ and $Z$ are d-separated from one another by a third set of nodes, $K$, then $P(q|z,k)=P(q|k)$. Thus if the value of each node in $K$ is known with certainty, a node in $Z$ cannot affect the computation for any node in $Q$.[1] This fact can be verified by an analysis of Pearl's equations for computation in a singly connected network.

2. The fact that a childless node for which no evidence is available does not affect the computation at any other node can be seen by examining Pearl's equations [Pearl 88] (pp 177-181) and the flow of information in the network. Figure 2 shows the information flow from a leaf, $x$. Information from $x$ that will eventually propagate to other nodes in the network, is sent to $x$'s immediate parents via a $\lambda$ parameter. $\lambda_x(u_i)$ denotes the vector that $x$ sends to it's parent, $u_i$. The vector has one element for each possible value of $u_i$.

The equation used to compute $\lambda$ parameters for each of node $x$'s parents is,

$$\lambda_x(u_i) = \beta \sum_x \lambda(x) \sum_{u_k: k \neq i} P(x|u) \prod_{k \neq i} \pi_x(u_k)$$

If $x$ is a childless node without evidence we set $\lambda(x)=(1,1,\cdots 1)$ and the equation above becomes,

$$\lambda_x(u_i) = \beta \sum_x \sum_{u_k: k \neq i} P(x|u) \prod_{k \neq i} \pi_x(u_k)$$

---

[1]The inference from sets of variables to individual variables is a simple application of the axiom of decomposition [Pearl 88].

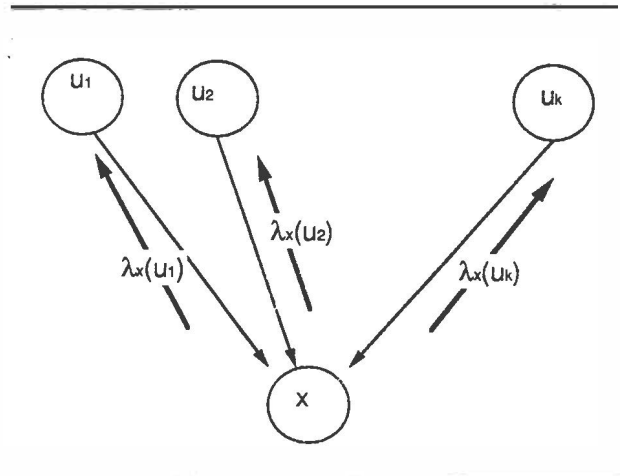

**Figure 2:** Propagation from the leaves

Each $\pi_x(u_k)$ is a vector with an entry for each value of $u_k$. The product, $\prod_{k \neq i} \pi_x(u_k)$, generates a matrix with a term for each combination of values for the parents, $u_k: k \neq i$. To see more easily what is going on, it is convenient to order these terms from 1 to $n$ and create a vector in which the $j^{th}$ entry, $\pi_j$, is a scalar equal to, $\pi(u_{1m})\pi(u_{2q})\cdots\pi(u_{kp})$, where $m$ represents one of the values of $u_1$, $q$ represents one of the values of $u_2$, etc. Similarly, $P(x|u,u_i)$ can be laid out in two dimensions which correspond to $x$ and the combinations of $u_{k \neq i}$. Holding $u_i$ constant, we can represent the product term in a matrix,

$$P(x|u,u_i)\prod_u \pi_x(u) = \begin{array}{c} x_1 \\ x_2 \\ \cdot \\ \cdot \\ x_m \end{array} \begin{bmatrix} u_1 \quad u_2 \quad \ldots\ldots \quad u_n \\ P(x_1|u_1,u_i)\pi \ldots\ldots\ldots P(x_1|u_n,u_i)\pi_n \\ \ldots\ldots\ldots\ldots\ldots\ldots\ldots\ldots\ldots \\ \ldots\ldots\ldots\ldots\ldots\ldots\ldots\ldots\ldots \\ P(x_m|u_1,u_i)\pi_1 \ldots\ldots\ldots P(x_m|u_n,u_i)\pi_n \end{bmatrix}$$

To compute $\lambda_x(u_i)$, we sum across both the rows and the columns of this matrix. Rearranging the order of the summation,

$$\beta \sum_x \sum_{j=1}^n P(x|u_j,u_i)\pi_j = \beta \sum_j \sum_x P(x|u_j,u_i)\pi_j$$

Examining the innermost sum and using the fact that $\sum_x P(x|U=u) =1$,

$$\sum_x P(x|u_j,u_i)\pi_j \;=\; \pi_j\sum_x P(x|u_j,u_i) \;=\; \pi_j$$

Finally, substitution into the original equation gives,

$$\lambda_x(u_i) \;=\; \beta\sum_j \pi_j \;=\; \beta\sum_j \pi(u_{1m})\pi(u_{2q})\cdots\pi(u_{kp})$$

where the $j^{th}$ combination of values for $u_{k\neq i}$ is,

$$u_{1m}u_{2q}\cdots u_{kp}.$$

The same argument holds for each value of $u_i$. This happens because the values of $u_i$ are reflected in the probabilities, $P(x|u_j,u_i)$, but these probabilities fail to affect the computation.

Since $\beta$ can be any convenient constant we select,

$$\beta \;=\; \frac{1}{\sum_j \pi_j} \quad \text{to get } \lambda_x(u_i)=(1,1\ldots\ldots1).$$

Thus, regardless of the information received from its parents, $x$, sends $\lambda_x(u_i)=(1,1,\cdots 1)$ to each of it's parents.

The result above depended only on the fact that the parameter, $\lambda(x)$, was set equal to $(1,1,\ldots1)$. As long as a node receives a $\lambda$ message from each of it's children that is a vector of ones, that node will have the property that information from it's parents will not affect outgoing $\lambda$ messages. This is a result of the fact that each node, $n$, that is neither a leaf nor in $K$ computes,

$$\lambda(n) \;=\; \prod_i \lambda_{x_i} \quad \text{where } x_i \text{ is a child of } n$$

Therefore, the computation at a node, $n$, does not affect the computation at any node reachable only on a path through the node's parents if the node's $\lambda(n)$ parameter is $(1,1,\ldots1)$. This is the case whenever the node receives a $\lambda$ parameter from each of it's children that is equal to $(1,1,\ldots1)$. Since a childless node without evidence, e.g., $x$, has it's $\lambda(x)$ parameter set to $(1,1\ldots1)$ and sends each of it's parents $\lambda$ parameters equal to $(1,1\ldots1)$, we can recursively remove these nodes from the network as long as they are not members of $Q$.

From the statement of the theorem it is not difficult to design an algorithm to construct computationally equivalent subgraphs. This is especially true because we can use **Algorithm 2** from [Geiger, Verma, and Pearl 89]. That algorithm takes as input a Bayesian network and two disjoint sets of nodes, e.g., $Q$ and $K$, and returns all nodes d-separated from $Q$ by $K$. The algorithm has $O(e)$ complexity, where $e$ is the number of edges in the Bayesian network. One way to use the algorithm in [Geiger, Verma, and Pearl 89] is to apply it to the Bayesian network to remove all the nodes that are d-separated from the query set, $Q$, and then to use depth-first search on the resulting subgraph to recursively prune any node that has no descendent in either $Q$ or $K$. A more efficient alternative is to alter their definition of a legal path so as to include paths which are not blocked but which have no descendents in either $Q$ or $K$.[2] Either method preserves the $O(e)$ complexity of their algorithm.

## 4 Minimal computationally equivalent subgraphs

We will now show that subgraphs generated by the method described in this paper are minimal in the sense intended by [Geiger, Verma, and Pearl 89]. That is, they are the best one can do using topological criteria alone. Smaller subgraphs may be possible. For example, it is conceivable that for a particular problem instance, a node could receive $\lambda$ parameters from each of its children whose quotient is a constant vector, e.g., (.75 2) and (1.25, .5). However, finding cases like this re-quires an examination of the probability distributions stored at the nodes in the network. The following definition formalizes the notion that determination of minimality of a subgraph should be independent of any of the probability distributions stored in the network.

---

[2]To do this, alter step (iii) of Algorithm 2 (p. 121) to read, "...2) v is not a head-to-head node on the trail u-v-w in D and $v\notin L$ *and descendent[v] = true.*" We should note that by using this modification to Algorithm 2, we can reduce the complexity of the computation that [Geiger, Verma, and Pearl 89] must use to achieve an equivalent result. This is due to the fact that they must double the number of nodes and introduce two edges for each new node in order to apply Algorithm 3.

**DEFINITION:** *A **minimal** computationally equivalent subgraph is a computationally equivalent subgraph such that for every node, $x \in Q$, in the subgraph, either,*

*1. there exists a node $q \in Q$ such that the computation at $q$ depends on the parameters of $x$, or*

*2. the fact that there is no $q \in Q$ that depends on the parameters of $x$ is dependent on the parameters of another node, $n \notin Q$, that is in the subgraph.*

According to this definition, a computationally equivalent subgraph cannot be minimal until each node is removed that is d-separated by $K = \{n \mid \text{the value of } n \text{ is known}\}$ from each of the nodes in $Q$. This is necessary because a nodes which are d-separated from $Q$ are, by definition, blocked from sending information to any node in $Q$. In addition, a subgraph is not minimal unless nodes which have no descendent in $Q \cup K$ are removed because these nodes prevent incoming information from passing to nodes reachable on undirected paths through their parents.

**Theorem 3.1:** *The computationally equivalent subgraph defined in Theorem 2.1 is minimal.*

**Proof:** The only way in which the probabilities stored at a node fail to affect the computation at other nodes reachable in the network is if they are d-separated from those nodes by a node with a known value or if another node along the path computes its $\lambda$ parameter to be a one vector. The probabilities stored at a node whose value is not known always affect the computation of the $\pi$ messages that are passed out of that node. This is a result of the fact that the $\pi$ parameter that $x$ sends to its child, $y_i$ is computed as,

$$\pi_{y_i} = \alpha \prod_{k \neq i} \lambda_{y_i}(x) \sum_u P(x|u)\pi_u$$

and, because neither $\lambda_{y_i}$ nor $\pi_u$ can be a *0* vector, the $\pi$ parameter is never independent of $x$'s probabilities, $P(x|u)$. We have already seen that the only way a node fails to pass its probabilities to its parents is if it computes its $\lambda$ message to be a one vector.

Because all the d-separated nodes have been removed, every node $x$ in the subgraph is on at least one active path to a node $q \in Q$. Therefore there is only one class of paths to consider: paths that reach $q$ through a node that computes its $\lambda$ parameter to be a one vector.

When a node, $x$, computes it's $\lambda$ parameter to be a one vector, that computation must be a result of either its own parameters or of the parameters of nodes downstream of $x$, i.e., through descendents of $x$ in the directed graph. This is the case because all paths in the subgraph that terminate in a leaf of the subgraph must terminate in a node in $Q \cup K$.

There are two cases: (1) paths in which the childless node is in $Q$ and (2) paths in which the childless node is in $K$. In the first case, there is an active path from $x$ to a $q$ through $x$'s children. This implies that either $x$'s $\pi$ messages to its children eventually affect the computation at $q$ or that path is blocked by a node that computes it's $\lambda$ vector to be one. In the second case, assume that the node, $t$, in $K$, takes on its $k^{th}$ value. It computes the $\lambda$ messages for its parents to be,

$$\lambda_t(u_i) = \beta \sum_j P(t_k|u_j, u_i)\pi_j$$

This is clearly not independent of $t$'s parameters. Thus the lambda messages computed at each ancestor, including $x$, will be a function of $t$'s parameters.

## 5 Conclusion

We have described a method by which a Bayesian network can be pruned at runtime without altering the computation at nodes whose values one is interested in. By preprocessing a network using this method it is possible to gain considerable savings in the amount of computation required to solve a given problem instance. Previous work [Geiger, Verma, and Pearl 89, Shachter 88], noted that similar results meant that it might be possible to reduce the number of parameters elicited from an expert when specifying a decision model. This is true when the problem instances that one wishes to solve never utilize some of the variables that make up the model. However, this is a condition that must hold over all problem instances encountered. If a problem instance arises that requires a set of parameters, those parameters must be elicited whether or not any other problem instance will ever utilize them. The benefits of network reduction for computational complexity can be applied to each problem instance independently.

The results described in this paper can be utilized for additional savings in a parallel distributed implementation. Under these conditions a preprocessing step need never be run. Instead, each node can use local information to determine whether its results will have an affect on the computations at other nodes. Whenever a node computes its own lambda vector to be a constant it need not send a message to any of its parents. As was pointed out in the previous section, lambda vectors may turn out to be constant as a function of the parameters stored at a node. Thus, savings beyond those achieved by a minimal subgraph can be realized in this type of implementation. Similarly, whenever a node's value is instantiated it may ignore any incoming information and thus (a) never recomputes it own value and (b) never needs update the messages it sends to its parents or children. Thus both the costs of communication and the cost of the vector multiplication in the receiving node is saved.

A limitation of the method that uses preprocessing to prune a network is that it may be inefficient when evidence is supplied incrementally. When the evidence is not provided all at once, nodes which had been pruned might have to be restored to the network. The simplest way to handle incremental evidence is to rerun the pruning algorithm on the original network each time an additional piece of evidence is encountered. If there are $v$ nodes in the Bayesian network, incremental evidence can result in a total cost of pruning of $O(ve)$. However, as long as the benefits in reduced computational complexity that comes from pruning leaf nodes outweighs this cost, the method described in this paper would still be an improvement. Moreover, it is likely that better techniques for restoring nodes to the subgraph can be developed so that the whole pruning algorithm need not be run each time an additional piece of evidence is encountered.